\documentclass[conference]{IEEEtran}
\IEEEoverridecommandlockouts
\usepackage{cite}
\usepackage{amsmath,amssymb,amsfonts}
\usepackage{algorithmic}
\usepackage{graphicx}
\usepackage{textcomp}
\usepackage{xcolor}
\usepackage{multirow}
\usepackage{booktabs}
\def\BibTeX{{\rm B\kern-.05em{\sc i\kern-.025em b}\kern-.08em
    T\kern-.1667em\lower.7ex\hbox{E}\kern-.125emX}}
\begin{document}

\title{Patch-aware Vector Quantized Codebook Learning for Unsupervised Visual Defect Detection}

\author{\IEEEauthorblockN{Qisen Cheng, Shuhui Qu, and Janghwan Lee}
\IEEEauthorblockA{\textit{Samsung Display America Lab} \\
San Jose, CA, USA, 95134 \\
qisen.c@samsung.com, shuhui.qu@samsung.com, jake.ee@samsung.com}

}

\maketitle

\begin{abstract}
 Unsupervised visual defect detection is essential across various industrial applications. Typically, this involves learning a representation space that captures only the features of normal data, and subsequently identifying defects by measuring deviations from this norm. However, balancing the expressiveness and compactness of this space is challenging. The space must be comprehensive enough to encapsulate all regular patterns of normal data, yet without becoming overly expressive, which leads to wasted computational and storage resources and may cause mode collapse—blurring the distinction between normal and defect data embeddings and impairing detection accuracy. To overcome these issues, we introduce a novel approach using an extended VQ-VAE framework optimized for unsupervised defect detection. Unlike traditional methods that apply a constant and uniform representation capacity across an image, our model employs a patch-aware dynamic code assignment scheme. This approach trains the model to allocate codes of varying resolutions based on the context richness of different image regions, aiming to optimize code usage spatially for each sample in a learnable fashion. We also leverage the learned strategy for code allocation in inference, to enlarge the discrepancy between normal and defective samples, thereby improving detection capabilities. Our extensive testing on the MVTecAD, BTAD, and MTSD datasets demonstrates that our model achieves state-of-the-art performance.
\end{abstract}


\section{Introduction}

Visual defect detection for quality control has been a significant focus in industry  \cite{Czimmermann2020,cheng202272}. With advancements in deep learning, numerous applications have emerged, such as detecting defective parts in electronics manufacturing \cite{balakrishnan20236,pan202432}, surface inspection in fabric or metal processing \cite{Mei2018, Ren2017}, dangerous events monitoring in traffic \cite{zhang2024research, li2024vehicle}. Unlike common scenarios where sufficient image samples are available for training, defect samples are often extremely rare or even nonexistent in production environments \cite{cheng202272, cheng2022estimation}. \par

\begin{figure}[htbp]
    \centerline{\includegraphics[width=3.5in]{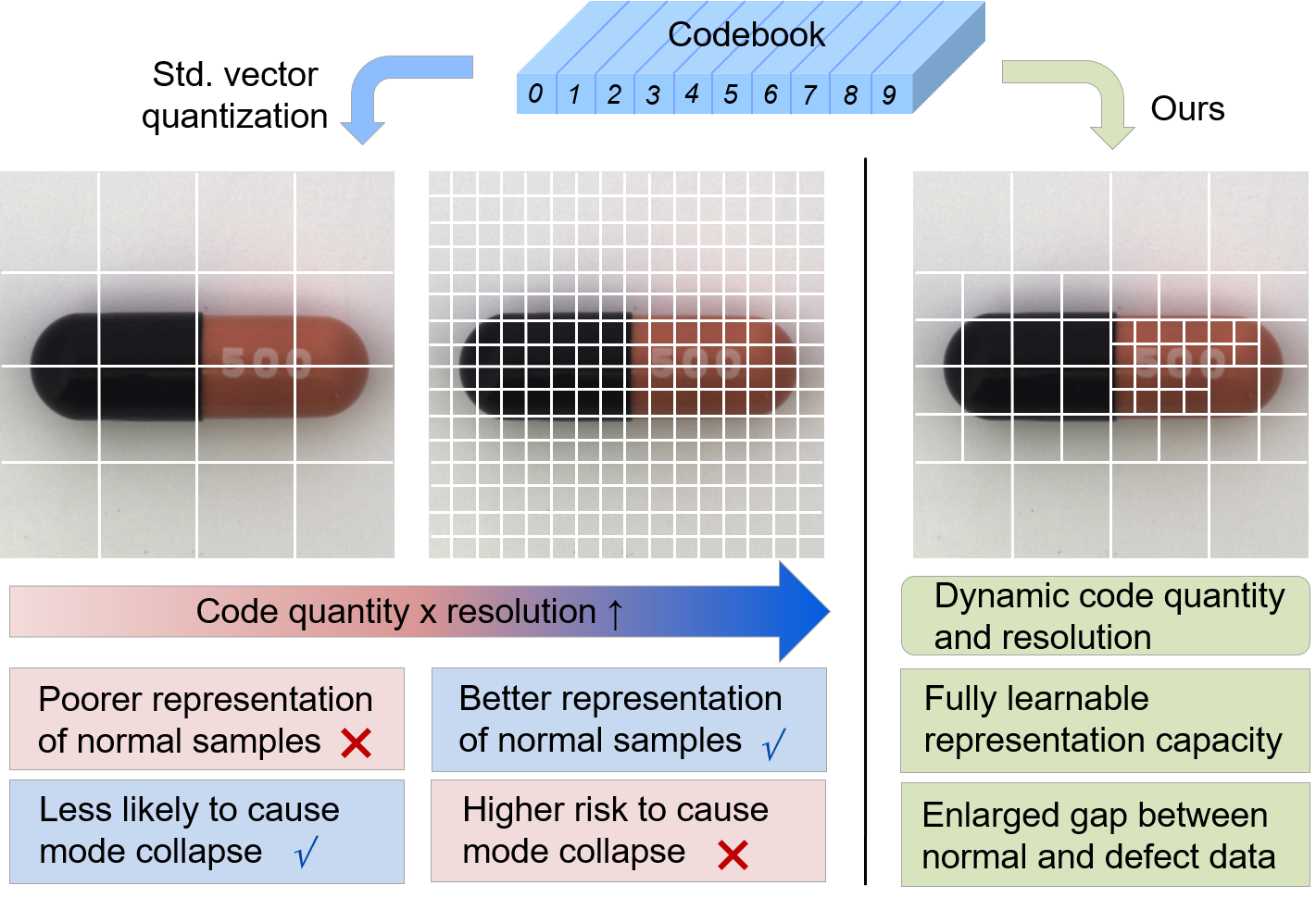}}
    \caption{Concept of unsupervised defect detection using patch-aware codebook learning. Visual patterns of the normal class are encoded into a discrete codebook; then the code is dynamically allocated to each sample with appropriate representation capacity, addressing the dilemma in prior works. The learned normal allocations are also leveraged for defect detection.}
    \label{fig:concept}
\end{figure}
To address the challenge of limited defect data, deep learning-based visual defect detection is often approached as an unsupervised learning task, known as one-class defect detection. This method assumes no defect data is available for training and instead focuses on creating a "memory of normality," which records the regular patterns of normal samples. Defects are then treated as anomalies that deviate from these memorized patterns, effectively turning defect detection into an outlier detection problem. The core idea is that the greater an input's deviation from established normal patterns, the more likely it is to be identified as a defect. Previous research has demonstrated that neural networks can significantly improve the memory of normality by generating detailed, compact representation spaces. A common approach involves training networks like auto-encoders to reconstruct normal samples, enabling the automatic learning of their representations \cite{liu2024deep}. However, this approach often encounters the "mode collapse" issue, where the model fails to differentiate between normal and defect samples during inference due to an overly expressive representation space \cite{chong2020simple}. An alternative strategy avoids training representation learning altogether, opting to use a pre-trained model as a feature extractor. These models store normal features in a memory bank, and defects are identified by measuring how much a sample's representation deviates from the stored features using methods such as k-nearest neighbors \cite{Defard2021, Roth2021}. However, these methods often require substantial memory capacity to capture the diversity of normal patterns, leading to higher computational and storage costs \cite{lu2023hierarchical}. Both strategies face the challenge of designing an effective representation learning mechanism, which involves a trade-off. On one hand, higher-capacity representation spaces can capture more detailed information about normal data. On the other hand, this increases the risk of mode collapse, where the model reconstructs inputs too accurately, regardless of whether they are normal or defective, thereby weakening its ability to detect anomalies. Recent approaches have sought to constrain the representation space to mitigate this issue. For example, MEMOAE uses a memory matrix for normality \cite{gong2019memorizing}, and vector quantization (VQ) methods like HVQ-Trans employ discrete codebooks rather than continuous spaces \cite{lu2023hierarchical}. While these methods show promise, they still require careful manual selection of representation capacity, which is difficult to optimize. Specifically, in vector quantization, coarse codes might overly restrict the latent space, leading to poor normal sample representation, whereas finer codes can improve normal sample representation but may introduce excessive expressiveness, code redundancy, and inefficiency.

To address the challenges in representation learning for unsupervised defect detection, we propose the Patch-aware Vector Quantized Autoencoder (PVQAE). This model enhances the conventional VQ-VAE framework by introducing a redesigned codebook learning mechanism, as shown in Figure \ref{fig:concept}. Typically, VQ assigns a fixed resolution and quantity of latent codes uniformly across an image, which limits the flexibility of representation. This uniform allocation does not account for the varying levels of detail across different regions of an image, resulting in suboptimal performance. Our approach solves this limitation by employing a dynamic code assignment scheme that adapts the resolution of codes based on the richness of context in each region. Finer codes are allocated to areas with detailed visual patterns, while coarser codes are assigned to regions with simpler, less informative patterns. This approach ensures efficient use of codes, allowing the model to reconstruct normal images with fewer, lower-resolution codes, while applying higher resolution only where necessary. During training, the model learns code budgets—referred to as \textit{normal budget priors}—which are fixed and used during testing. These priors prevent the model from utilizing excessive reconstruction capabilities, particularly in regions with unexpected detailed patterns, such as defects. For instance, a crack with sharp, irregular edges on the smooth surface of a glass cup would require more detailed reconstruction in those unexpected areas. By limiting the model's ability to reconstruct such patterns, the resulting higher reconstruction error makes defect detection more effective. Moreover, the flexibility in allocating latent codebook capacity allows our PVQAE model to handle a wide range of visual patterns across multiple objects with a single model. This eliminates the need for training separate models for each product type, which has been a labor-intensive practice in prior work \cite{Defard2021, Roth2021}.

\begin{figure*}[htbp]
    \centerline{\includegraphics[width=\textwidth]{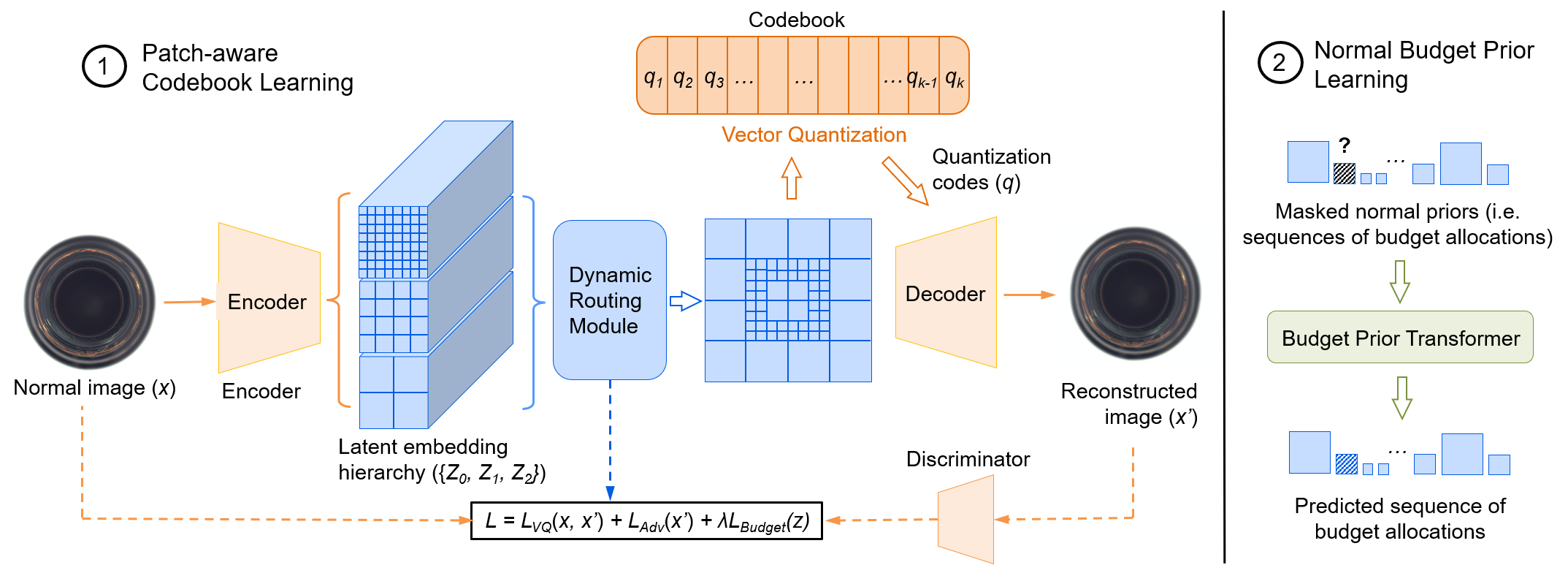}}
    \caption{Visualization of PVQAE model. It extends the vector quantization technique with 1) patch-aware codebook learning to optimize the representation capacity, i.e. budget, for each sample, and 2) normal budget prior learning step to record the regular code usage patterns.}
    \label{fig:pvqae}
\end{figure*}
 
Our method achieved superior performance in both image and pixel-level evaluations on several defect detection benchmarks, compared to existing state-of-the-art baselines. These experiments validate the effectiveness of our method. To summarize, this paper makes three key contributions:  \par 
1. To the best of our knowledge, we are the first to explore how a one-class model can find optimal representation capacity during training, addressing a key challenge in defect detection. \par
2. We introduced two novel approaches to enhance unsupervised defect detection performance: patch-aware dynamic code allocation and normal budget prior learning. \par
3. Our PVQAE model demonstrated superior performance across multiple industrial defect detection datasets.

The rest of the paper is structured as follows: Section 2 covers related work, Section 3 details our codebook learning and detection method, Section 4 presents experimental results, and Section 5 concludes the paper.

\section{Related work}


\subsection{Unsupervised Defect Detection}



Most existing work uses a one-class approach, relying on learning representations of normal data and identifying defects as outliers. Early methods used minimal-volume spheres \cite{Deng2009} or predefined kernels \cite{Ruff2018} to simplify the representation space, which are later improved by adding geometric transformations \cite{Golan2018}, patch-wise embeddings \cite{Yi2020}, hybrid model architectures \cite{li2024enhancing} and contrastive learning \cite{li2021cutpaste}. Generative models like autoencoders \cite{Sakurada2014, kang2019} and GANs \cite{Sabokrou2018} have also been used for more complex data. Some other techniques introduced attention-guided models \cite{Venkataramanan2020}, Gaussian Mixture Models \cite{Zong2018}, normalizing flows \cite{Rudolph2021, gudovskiy2021cflow}, and knowledge distillation \cite{Bergmann2020, wang2021student} to learn condensed representations. Some recent works avoid training altogether by leveraging pre-trained models \cite{cohen2020} with local distribution fitting \cite{Defard2021} and Coreset sampling \cite{Roth2021}. However, the heavy reliance on pre-trained datasets may be sub-optimal for domain-specific tasks like industrial defect detection. \par

\subsection{Vector Quantization}
Vector quantization (VQ), introduced in the 1980s for signal compression \cite{gersho2012vector}, has later been combined with deep generative models like VQ-VAE \cite{van2017neural, yunoki2023exploring}, VQ-GAN \cite{esser2021taming} and LLM \cite{li2024integrated, saunders2024optimizing}. Standard VQ encodes input data as a codebook of discrete codes, enabling expressive representations while preventing mode collapse. This approach has inspired unsupervised defect detection methods \cite{lu2023hierarchical, kim2024patch}. PVQAE also uses VQ but improves on it by introducing a dynamic codebook learning process with adjustable code capacity, addressing the limitations of fixed codebooks for defect detection.


\section{Methodology}

Our PVQAE model is illustrated in Figure 2. It extends the VQ-VAE framework with a patch-aware dynamic codebook learning scheme and a normal budget prior learning step, and is trained in an end-to-end fashion. In this section, we will first revisit the standard VQ-VAE framework and then describe our proposed method in detail.

\subsection{Preliminary}
\subsubsection{Vector Quantized Variational AutoEncoder (VQ-VAE)}
A VQ-VAE model consists of a pair of encoder ($E$) and decoder ($G$), and a latent representation structure. The latent representation structure is formulated as a codebook $Q \in \mathbb{R}^{(K \times n)}$, which contains $K$ number of discrete codes (i.e. embeddings) $q_k$, $k \in 1,2,…,K$, each of dimension $n$. To learn the codebook, an input image $x$ is first encoded by $E$ to a matrix of discrete embeddings $Z \in \mathbb{R}^{(h \times w \times n)}$, where each embedding $z_{i,j}$ correspond to one image constituent (i.e. patch). Then for each image patch, a code $q_{i,j}$ is assigned to it by using a look-up function $l(\cdot)$, which selects a certain code from the codebook $Q$ that has the smallest euclidean distance from patch embedding $z_{i,j}$:


\begin{center}
    ${q}_{i,j}=l(z_{i,j}, Q)=argmin_{m}\left \| z_{i,j}-q_{m} \right \|_{2}^{2}$\\
\end{center}\par
The matrix of assigned codes, denoted as $q$, is then passed to the decoder network to generate a reconstructed image ($\hat{x}$):
\begin{center}
    $\hat{x}=G(z_{q})=G(q(E(x)))$
\end{center}\par
The model is trained with 2 main objectives: 1) reconstruction loss measuring the difference between reconstructed image and the input, and 2) discrete representation learning losses encouraging the alignment between the selected latent codes and the encoder outputs. Since the $argmin$ operand used in the look-up function is not differentiable during backpropagation, the gradient of this step is approximated by using the straight-through estimator (STE) with stop-gradient operand ($sg$) \cite{van2017}. The overall loss is summation of the objectives with $\beta$ as a hyper-parameter:
\begin{center}
$\mathcal{L}_{VQ}(E,G,Q))=\left \| \hat{x}-x \right \|_{2}^{2}+\left \| sg[E(x)]-q \right \|_{2}^{2}+\beta\left \| sg[q]-E(x) \right \|_{2}^{2}$
\end{center}

\subsubsection{Adversarial Training}
Adversarial training also benefits representation learning as reported in \cite{esser2021taming}. We hence incorporate it enhance the representations to fully express regular patterns. Specifically, we add a patch-wise discriminator network $D$, and introduce an additional classification loss to distinguish between reconstructed and real patches:
\begin{center}
$\mathcal{L}_{ADV}((E,G,Q),D)=log(D(x))+log(1-D(\hat{x}))$
\end{center}

\subsection{Patch-aware Vector Quantized Codebook Learning}
\subsubsection{Multi-resolution dynamic code allocation}
Existing methods typically use a codebook with a single static resolution \cite{kim2024patch} or a combination of static resolutions \cite{lu2023hierarchical}, assigning a constant number of latent codes to an image. In contrast, our PVQAE dynamically assigns latent codes of varying resolutions and quantities. We define a hierarchy of candidate resolutions \( R = \{r_1 < r_2 < ... < r_k\} \), where \( r_i \in (0, 1) \), ordered by the area of the image each resolution represents. For each resolution \( r \), the encoder \( E \) encodes an input image \( I \in \mathbb{R}^{H \times W \times 3} \) into a matrix of feature embeddings \( Z_r = \{{z_r}^{1,1},{z_r}^{1,2}, ..., {z_r}^{1/r, 1/r}\} \), where each \( {z_r}^{i,j} \in \mathbb{R}^{(H \times r) \times (W \times r) \times d} \) corresponds to an image patch \( I^{i,j} \). This creates a hierarchy of embeddings at multiple resolutions \( Z = \{Z_1, Z_2, ..., Z_R\} \).

\begin{figure}[htbp]
    \centerline{\includegraphics[width=3.5in]{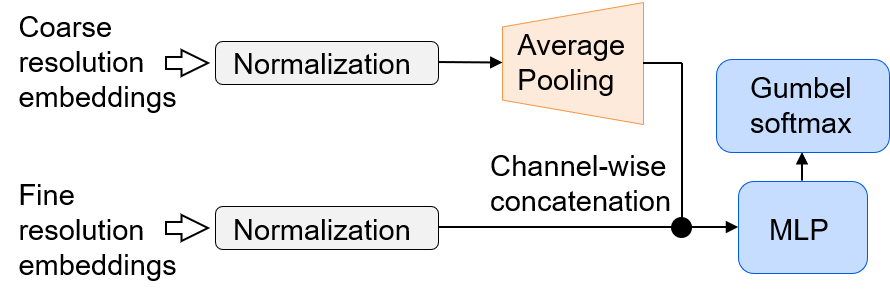}}
    \caption{Detailed architecture of \textit{Dynamic Routing Module}.}
    \label{fig:drm}
\end{figure}

Recognizing that different regions of an image contain varying levels of contextual information, we select the most appropriate resolution for each region. Inspired by \cite{huang2023towards}, we use a \textit{Dynamic Routing Module} to determine the optimal resolution. As shown in \ref{fig:drm}, the module applies average pooling to feature embeddings at different resolutions, except for the finest level, ensuring all embeddings are resized to match the dimensions of \( (H \times r_1) \times (W \times r_1) \times d \). These pooled embeddings are concatenated and passed through an \textit{MLP} layer, which acts as a gating mechanism, outputting logits \( g^{i,j} \in \mathbb{R}^R \) for the resolution levels of each image patch \( I^{i,j} \) at the coarsest resolution. The Gumbel-Softmax technique \cite{jang2016categorical} is applied to these logits, providing a soft, differentiable approximation of the \( argmax \) operation. For each patch \( I^{i,j} \), a score \( b \) is generated for the resolution levels:

\begin{center}
${b_{r}}^{i,j}=\frac{exp({g_r}^{i,j} + {\delta_r}^{i,j})/\tau}{\sum_{r=1}^{R}{exp({g_r}^{i,j} + {\delta_r}^{i,j})/\tau}}$
\end{center}
\begin{center}
$\sum_{r=1}^{R}{b_{r}^{i,j}}=1$
\end{center}
where $\delta$ is random noise sampled from Gumbel distribution, and $\tau > 0$ is the temperature adjusting the sharpness of scoring. As the temperature approaches 0, the score becomes closer to an one-hot hard score. At the selected resolution level, discrete latent codes are assigned to the feature embeddings as in the standard vector quantization. Consequently, each image are represented with codes of different length.

\subsubsection{Progressive budget learning}

Our goal is to ensure the model represents each sample with the most economical representation capacity. This means assigning fine-resolution codes to regions with rich context and coarse-resolution codes to areas with simple patterns, such as smooth product surfaces with uniform textures. To achieve this, we introduce a budget loss to guide resolution selection, penalizing the use of finer codes by assigning them a higher cost. Specifically, we apply discrete wavelet transformation (DWT) to each image patch at the coarsest resolution level and compute the normalized entropy (\(\mathcal{H}\)) of the DWT coefficients to estimate context richness. The inverse of \(\mathcal{H}\) is then used as the base cost, making it context-dependent—cheaper for regions with richer context. As code resolution increases, the cost rises by a factor of \(c > 1\), i.e. \(c = 2^{R-1}\). The overall budget loss is the summation of costs across all patches \(I^{i,j}\) at the coarsest level:

\begin{center}
$\mathcal{L}_{Budget} = \sum_{i,j=1}^{i=H/r_k, j=W/r_k}{\frac{1}{\mathcal{H}(I^{i,j})}  c^{r}}$
\end{center}
\begin{center}
$\Tilde{\mathcal{H}}(I^{i,j}) = \frac{\mathcal{H}(DWT(I^{i,j}))}{\sum_{i,j}{\mathcal{H}(DWT(I^{i,j}))}} $
\end{center}
Then the overall training objective is:
\begin{center}
$\mathcal{L} = \mathcal{L}_{VQ} + \mathcal{L}_{ADV} + \lambda \mathcal{L}_{Budget} $
\end{center}
where $\lambda$ is a hyper-parameter to adjust the weight on budget learning. To prevent the model falls to a situation where it avoids using fine resolution codes at all, we schedule $\lambda$ to linearly increase from 0 to the maximum value. This encourages the model to learn expressive latent representations and proper reconstruction of normal images first, and then progressively pivot towards refining the representation capacity.

\subsection{Defect Detection with PVQAE}
\subsubsection{Learning Normal Budget Priors}
The learned budget represents the allocation of code resolutions in different regions, capturing the regular distribution of normal patterns. We leverage it as an additional indicator of normality for defect detection. For instance, in the MVTecAD dataset, normal images of drug capsules likely have finer resolution codes to regions with intricate details, e.g. the red half shell with white printings. During inference, if more budget is needed for unexpected regions, or less to the detailed areas, it may indicate defects. To record the learned budget priors, we first flatten the resolution level matrix into a sequence in a clockwise order: \( B_{norm} = \{ b_{r}^{1,1}, b_{r}^{1,2}, ..., b_{r}^{H/r_k, W/r_k} \} \). Inspired by VQ-GAN \cite{esser2021taming}, we then train a standalone \textit{Budget Prior Transformer} to capture these normal sequences. The task is framed as predicting a masked token (resolution level) within the sequence by referencing all other tokens, enumerating the masks to obtain the fully predicted sequence \(\hat{B}_{norm}\). This approach relies on the observation that defects are typically localized, while most image regions remain normal, meaning the normal budget for a region can be inferred from the surrounding areas. To accommodate object-class-specific budget learning, we prepend a \textit{CLS} token to \(B_{norm}\). During training, we freeze the rest of the model to obtain \(B_{norm}\) from the \textit{Dynamic Routing Module} and train the \textit{Budget Prior Transformer} using Cross Entropy (CE) loss:
\begin{center}
$\mathcal{L}_{Prior} = \sum_{i,j}{CE(b_{norm}^{i,j}, \hat{b}_{norm}^{i,j})} $
\end{center}

\subsubsection{Scoring for Defect Detection}
Given an input image \( x \), PVQAE detects defects by estimating a defect score \( S \) for each pixel. This score consists of two components: \( \mathcal{S}_{Prior} \) and \( \mathcal{S}_{Recon} \). \( \mathcal{S}_{Prior} \) measures the Cross Entropy between the budget \( B \), dynamically determined by the \textit{Dynamic Routing Module} based on the input image's context, and the normal budget prior \( \hat{B}_{norm} \), predicted by the \textit{Budget Prior Transformer}. \( \mathcal{S}_{Recon} \) is the L2 reconstruction error. During reconstruction, we use \( \hat{B}_{norm} \) for code allocation to prevent the model from utilizing excessive representation power, thereby limiting the reconstruction quality for unseen defects. The final defect score is the pixel-wise product of the two components, which is thresholded for the eventual detection:
\begin{center}
$\mathcal{S} = \mathcal{S}_{Prior} \times \mathcal{S}_{Recon}$, where
\end{center}

\begin{center}
$\mathcal{S}_{Prior} = \frac{CE(b^{i,j}, \hat{b}_{norm}^{i,j})}{\sum_{i,j}{CE(b^{i,j}, \hat{b}_{norm}^{i,j})}}$, $\mathcal{S}_{Recon} = \left \| \hat{x}-x \right \|_{2}^{2}$
\end{center}


\begin{center}
$\mathbb{I}(s_{i,j}>t)\;\forall\; s_{i,j}\in S$
\end{center} \par

\section{Experiments}

In this section, we present the experimental evaluation of our PVQAE model on public manufacturing datasets for both image and pixel-level defect detection. We compare our model with several popular baselines. For memory-based methods, we include the latest approaches, PaDiM \cite{Defard2021} and PatchCore \cite{Roth2021}. Among reconstruction-based methods, we focus on AE-based techniques such as the AE-SSIM \cite{Sakurada2014}, VQ-E \cite{kim2024patch} (designed for medical image anomaly detection), and HVQ-Trans \cite{lu2023hierarchical}, a recent state-of-the-art defect detection model. The latter two use standard VQ techniques with static resolution codebooks. We also include STPM \cite{wang2021student} and CutPaste \cite{li2021cutpaste} for a comprehensive comparison across different representation learning methods. Following \cite{Roth2021}, the area under the receiver-operator curve (AUROC) is used as the evaluation metric. Additionally, we qualitatively assess the effectiveness of our dynamic code allocation for defect detection and conclude with ablation studies on progressive budget learning and normal budget priors.

\subsection{Implementation Details \& Datasets}
We built our PVQAE on VQ-GAN \cite{esser2021taming}, using its encoder and detector configurations. Outputs from the last three encoder convolutional blocks formed the hierarchy of feature embeddings. The \textit{Dynamic Routing Module} followed \cite{huang2023towards}, with an \textit{MLP} layer containing a single hidden layer to reduce computation. The \textit{Budget Prior Transformer} used one transformer block with a learnable embedding matrix $E \in \mathbb{R}^{3\times d}$ to tokenize code resolution indices. Images were resized to 256$\times$256 and augmented with random flipping and color-jittering. A single model was trained across all objects in a dataset. Training was conducted on an Intel Core i9-9920X CPU and NVidia RTX8000 48G GPU. Public implementations from \cite{wang2024real} and \cite{lu2023hierarchical} were used for baselines, excluding the optimal-transport objective from HVQ-Trans to focus on representation learning comparison. \par
We evaluated on three datasets: 1) MVTecAD \cite{Bergmann2019} with 5354 images in 15 classes; 2) BTAD \cite{btad} with 2830 images from 3 products, 3) MTSD \cite{Huang2020} with 925 normal and 392 defective tile images. BTAD and MSTD are combined in training due to limited data size.

\subsection{Quantitative Defect Detection Performance}
\subsubsection{MVTecAD}
The results for MVTecAD are presented in Table \ref{tab:perf}. Our PVQAE method demonstrates competitive performance compared to all other baselines in both image and pixel-level AUROC. Notably, with 15 different product categories, it was challenging to learn a single latent representation that adequately captured the diverse distribution. For fairness, we trained and tested all methods on multiple objects simultaneously, enabling us to highlight differences in representation learning capability. While the SOTA memory-based method, PatchCore, produced strong results across several objects, it was burdened by high computation and storage costs due to the need to cover multiple objects. In comparison, our PVQAE method significantly outperformed VQ-E, which relies on standard vector quantization, across all objects. HVQ-Trans, a recent strong baseline, also achieved competitive scores but was considerably more resource-intensive due to its Mixture-of-Expert mechanism that switches between multiple codebooks for different objects, making it relatively less efficient.

\begin{table*}[htbp]
    \caption{Quantitative Defect Detection Performance. Image-level AUROC \% / Pixel-level AUROC \%. }
\begin{center}
\begin{tabular}{|c|c|c|c|c|c|c|c|c|c|}
\hline
\multirow{3}{*}{\textbf{Dataset}} & \multirow{3}{*}{\textbf{Category}} & \multicolumn{8}{|c|}{\textbf{Model}} \\
\cline{3-10}
 & & \multicolumn{2}{|c|}{\textbf{Memory-based}} & \multicolumn{3}{|c|}{\textbf{Reconstruction-based}} & \multicolumn{2}{|c|}{\textbf{Others}} & \multirow{2}{*}{\textbf{\textit{Ours}}} \\
\cline{3-9} 
 & & \textbf{\textit{PaDiM}} & \textbf{\textit{Patchcore}} & \textbf{\textit{AE-SSIM}} &  \textbf{\textit{VQ-E}} & \textbf{\textit{HVQ-Trans}} & \textbf{\textit{STPM}} & \textbf{\textit{CutPaste}} & \\ 
\cline{1-10} 
\multirow{15}{*}{MVTecAD} & Carpet
        & 91.9 / 92.7 & 96.8 / \textbf{98.7} & 81.2 / 82.4 & 90.4 / 93.5 & 96.7 / 97.4 & 91.2 / 89.5 & 89.2 / 90.3 & \textbf{98.4} / 95.2 \\
\cline{2-10}
                          & Grid
        & 73.2 / 70.8 & 90.5 / 97.6 & 70.5 / 69.1 & 83.6 / 87.3 & 96.4 / \textbf{96.8} & 89.7 / 90.3 & 88.2 / 89.3 & \textbf{97.6} / 94.2 \\
\cline{2-10}
                          & Leather
        & 98.8 / 99.2 & \textbf{100} / 98.2 & 67.2 / 74.3 & 91.4 / 93.1 & \textbf{99.2} / 97.3 & 92.2 / 89.1 & 91.2 / 93.5 & 99.4 / 97.5 \\
\cline{2-10}
                          & Tile
        & 93.1 / 81.3 & 92.3 / 90.2 & 77.2 / 69.4 & 87.2 / 90.3 & \textbf{98.5} / 91.4 & 90.8 / 86.3 & 92.2 / 93.5 & 95.4 / \textbf{93.6} \\
\cline{2-10}
                          & Wood
        & \textbf{98.4} / 87.3 & 91.6 / 90.2 & 65.6 / 72.1 & 91.4 / 90.5 & 96.6 / \textbf{93.7} & 89.7 / 90.3 & 95.2 / 89.3 & 97.2 / 91.3 \\
\cline{2-10}
                          & Bottle
        & 96.9 / 94.1 & \textbf{100} / 95.2 & 82.2 / 86.1 & 91.7 / 89.2 & \textbf{100} / 97.9 & 99.2 / 96.1 & 97.2 / 97.7 & \textbf{100} / \textbf{98.1} \\
\cline{2-10}
                          & Cable
        & 73.2 / 78.9 & 93.6 / 91.3 & 67.2 / 61.6 & 89.5 / 88.3 & \textbf{99.1} / \textbf{98.0} & 95.2 / 94.7 & 92.8 / 93.2 & 95.3 / 96.9 \\
\cline{2-10}
                          & Capsule
        & 75.6 / 93.2 & \textbf{97.5} / 97.1 & 81.2 / 73.3 & 93.6 / 92.3 & 97.3 / 97.8 & 95.2 / 96.4 & 96.8 / 92.1 & 97.2 / \textbf{98.5} \\
\cline{2-10}
                          & Hazelnut
        & 86.6 / 94.9 & 99.3 / 96.5 & 81.7 / 76.1 & 92.5 / 88.1 & \textbf{100} / \textbf{98.4} & 94.2 / 93.8 & 96.2 / 97.3 & \textbf{100} / 98.1 \\
\cline{2-10}
                          & Metal Nut
        & 85.7 / 88.9 & 97.6 / 97.1 & 61.2 / 70.5 & 91.3 / 88.4 & \textbf{100} / 97.4 & 96.2 / 95.0 & 95.2 / 91.8 & 99.4 / \textbf{97.8} \\
\cline{2-10}
                          & Pill
        & 69.7 / 86.9 & 97.3 / 96.5 & 79.2 / 72.1 & 88.4 / 91.2 & 96.2 / 95.8 & \textbf{96.6} / 97.2 & 95.2 / 94.1 & 95.7 / \textbf{97.4} \\
\cline{2-10}
                          & Screw
        & 55.8 / 87.3 & 90.3 / \textbf{98.4} & 54.2 / 49.9 & 87.0 / 89.7 & 93.2 / 90.5 & 91.5 / 92.8 & \textbf{95.4} / 91.8 & 92.6 / 93.1 \\
\cline{2-10}
                          & Toothbrush
        & 94.8 / 94.8 & \textbf{99.8} / \textbf{98.3} & 77.2 / 58.1 & 79.4 / 85.5 & 91.0 / 96.3 & 93.8 / 92.1 & 92.4 / 96.5 & 93.7 / 95.8 \\
\cline{2-10}
                          & Transistor
        & 87.8 / 91.7 & 98.5 / 91.8 & 62.4 / 47.7 & 91.5 / 95.7 & 99.2 / \textbf{96.1} & 71.3 / 80.5 & 84.6 / 88.9 & \textbf{99.7} / 94.3 \\
\cline{2-10}
                          & Zipper
        & 77.2 / 93.8 & 97.3 / 95.1 & 77.2 / 73.2 & 85.4 / 87.6 & 95.8 / 96.6 & 95.7 / \textbf{97.2} & 96.8 / 95.6 & \textbf{98.4} / 96.7 \\
\hline
\multirow{3}{*}{BTAD} & Prod 1
        & 97.2 / 95.3 & \textbf{100} / 97.9 & 89.8 / 85.7 & 95.6 / 94.1 & \textbf{100} / 98.4 & 99.6 / 95.8 & 99.0 / 97.3 & \textbf{100} / \textbf{98.9} \\
\cline{2-10}
                      & Prod 2
        & 98.1 / 93.4 & \textbf{100} / 97.9 & 86.2 / 82.2 & 99.6 / 97.1 & \textbf{100} / 98.2 & \textbf{100} / 97.3 & \textbf{100} / 96.9 & \textbf{100} / \textbf{98.5} \\
\cline{2-10}
                      & Prod 3
        & 95.6 / 97.2 & 99.2 / \textbf{97.6} & 80.5 / 82.4 & 97.0 / 96.5 & 99.2 / 97.5 & \textbf{100} / 92.7 & 99.2 / 97.3 & 99.4 / \textbf{97.6} \\
\hline
\multirow{1}{*}{MTSD} & Magnets
        & 99.2 / 93.2 & 99.2 / 94.6 & 89.2 / 87.1 & 96.6 / 94.2 & \textbf{100} / 94.5 & 90.2 / 91.4 & \textbf{100} / 90.7 & 99.2 / \textbf{96.5} \\
\hline
\end{tabular}
\label{tab:perf}
\end{center}
\end{table*}

\subsubsection{BTAD \& MTSD}
We combined BTAD and MTSD datasets together in training for all methods. Similarly, our PVQAE method outperformed or achieved comparative performance on the BTAD and MTSD datasets by comparing to all other baselines. The superior performance of PVQAE was probably because our codebook learning not only produced an expressive codebook enabling good coverage of normal visual patterns, but also avoided a over-expressive latent space that could lead to better reconstruction of the defects. \par


\begin{figure}[htbp]
    \centerline{\includegraphics[width=3.5in]{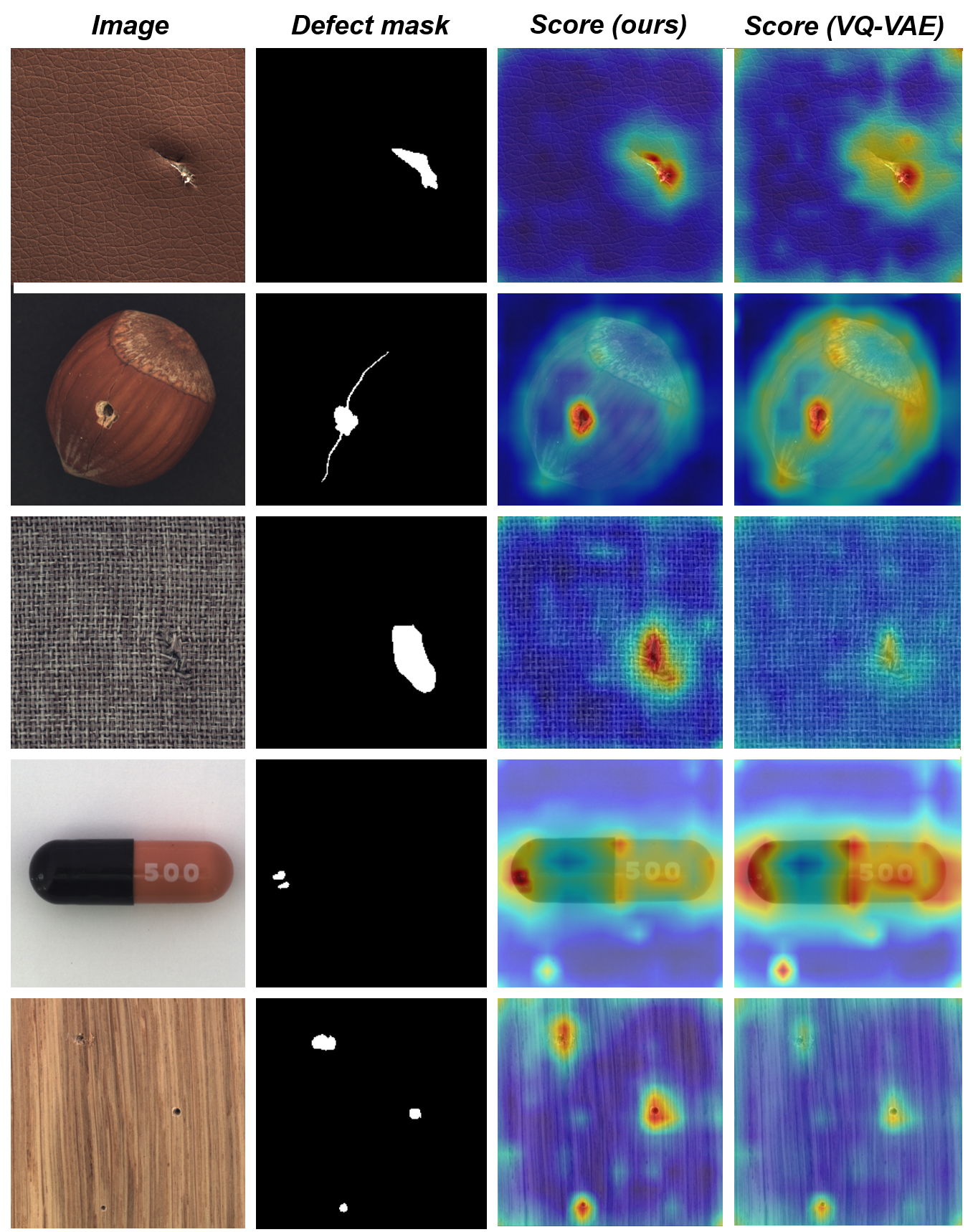}}
    \caption{Pixel-level defect detection samples on MVTecAD dataset.}
    \label{fig:pixel_perf}
\end{figure}

\subsection{Qualitative Defect Detection Performance}

In this subsection, we qualitatively investigated the effectiveness of our method by comparing pixel-level defect detection between the standard VQ-VAE and our PVQAE. Figure \ref{fig:pixel_perf} shows the defect detection results on multiple objects. To highlight our method's advantage, we focused on defect samples where the standard VQ-VAE struggled to locate and segment defective areas, particularly when defects were small or subtle. The defect scores from both methods are displayed side by side, with pixel-level scores visualized using a color spectrum—hotter colors (e.g., red) indicating higher defect likelihood and cooler colors (e.g., blue) representing lower likelihood. It is evident that the score maps from PVQAE had more concentrated red areas precisely aligning with the defect regions and cleaner backgrounds in normal areas. This demonstrates that our dynamic code allocation and learned normal budgets enhanced the contrast between defects and normal samples, significantly improving detection performance.

\subsection{Ablation Analysis}
We further analyzed the effectiveness of the proposed budget learning mechanism through two ablation studies: one on the progressive budget learning scheme and another on defect detection using learned normal budget priors.

\subsubsection{Progressive Budget Learning}
\begin{figure}[htbp]
    \centerline{\includegraphics[width=3in]{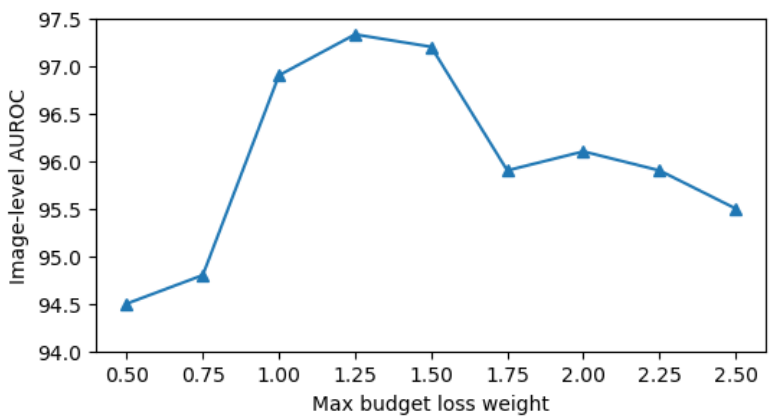}}
    \caption{Defect detection performance on MVTecAD vs. budget loss weight}
    \label{fig:maxweight_ablation}
\end{figure}

Progressive budget learning is controlled by the hyper-parameter \(\lambda\), which adjusts the weight of the budget learning loss in the overall objective. To understand the impact of \(\lambda\), we conducted two ablation studies on its scheduling and the terminal maximum weight. First, we tested three different \(\lambda\) schedules: 1) \textit{Constant}=1, 2) \textit{Cosine} schedule varying \(\lambda\) from 0 to 1, and 3) \textit{Linear} schedule with \(\lambda\) from 0 to 1. As shown in Table \ref{tab:schedule_ablation}, the \textit{Linear} schedule provided the best defect detection performance. Second, we examined the optimal choice for maximum \(\lambda\) by using a linear schedule with varying terminal weights, ranging from 0.5 to 2.5 in steps of 0.25. As shown in Figure \ref{fig:maxweight_ablation}, detection performance improved as the weight increased, peaking around 1.25, after which performance declined with further increases.

\subsubsection{Normal Budget Priors}
We also evaluated the effectiveness of normal budget priors in defect detection. In this study, we first tested removing the prior learning but relying on dynamically allocated code budgets during inference. We then compared this with models using either universal and per-class priors. The only difference was whether a class token was added to budget sequence during training. We reported the average image and pixel-level AUROCs on MVTecAD for comparison. As shown in Table \ref{tab:prior_ablation}, per-class prior achieved the best performance, demonstrating its effectiveness.

\begin{table}[]
    \caption{Ablation study of weight schedule on MVTecAD dataset}
    \vspace{0.5ex}
    \centering
    \begin{tabular}{l l l l}
    \toprule
              & Constant & Cosine schedule & Linear schedule \\
        \midrule
        Image-AUROC (mean) & 89.2 & 96.3 & 96.9 \\
        \midrule
        Pixel-AUROC (mean) & 87.7 & 92.5 & 94.2 \\
    \bottomrule
    \end{tabular}
    \label{tab:schedule_ablation}
\end{table}

\begin{table}[]
    \caption{Ablation study of normal budget priors on MVTecAD dataset}
    \vspace{0.5ex}
    \centering
    \begin{tabular}{l l l l}
    \toprule
              & No priors & Priors & Per-class priors \\
        \midrule
        Image-AUROC (mean) & 91.5 & 96.2 & 97.3 \\
        \midrule
        Pixel-AUROC (mean) & 90.8 & 93.1 & 95.9 \\
    \bottomrule
    \end{tabular}
    \label{tab:prior_ablation}
\end{table}

\section{Conclusion}
In this paper, we present the Patch-aware Vector Quantized Auto-Encoder (PVQAE), a refined model for unsupervised defect detection. It improves enhances the conventional VQ-VAE by specifically tailoring the codebook learning process for one-class defect detection. Unlike previous methods, our approach dynamically allocates finer latent codes to regions with rich context and coarser ones to less informative areas. This data-driven method removes the need for manual capacity setting. A budget learning objective enables optimal code allocation end-to-end, making the capacity fully adaptable. Normal sample priors further enhance defect detection by constraining defect reconstruction quality. Our method offers key insights for one-class defect detection systems.

\bibliographystyle{ieeetr}
\bibliography{conference_101719}


\end{document}